\documentclass{bmvc2k}
\usepackage{amsmath,amsfonts,amssymb}
\usepackage{commath}
\usepackage{mathtools}
\usepackage{multicol}
\usepackage{graphicx}
\usepackage{float}
\usepackage{amsmath}
\usepackage{algorithm}
\usepackage{algpseudocode}
\usepackage[dvipsnames]{xcolor}
\usepackage{multirow}
\usepackage{colortbl}
\usepackage{pst-node}
\usepackage{pdftricks}
\usepackage{pst-plot}
\usepackage{pst-sigsys}
\usepackage{pstricks-add}
\usepackage{booktabs} 

\newcommand{\mypar}[1]{{\bf #1.}}

\definecolor{darkredc}{RGB}{90,0,0}
\definecolor{darkgreen}{RGB}{0,90,0}
\definecolor{darkblue}{RGB}{0,0,90}			
\definecolor{thisblue}{rgb}{0,0,.4}
\definecolor{myred}{RGB}{153,0,0}
\definecolor{myblue}{RGB}{0,0,153}

\title{Single Image Super-Resolution via CNN Architectures and TV-TV Minimization}

\addauthor{Marija Vella}{mv37@hw.ac.uk}{1}
\addauthor{Jo\~{a}o F. C. Mota}{j.mota@hw.ac.uk}{1}

\addinstitution{
 Institute of Sensors, Signals and Systems\\
 Heriot-Watt University\\
 Edinburgh, UK
}

\runninghead{Vella, Mota}{SR via CNN Architectures and TV-TV Minimization}

\begin{document}

\maketitle

\begin{abstract}

Super-resolution (SR) is a technique that allows increasing the resolution of a given image. Having applications in many areas, from medical imaging to consumer electronics, several SR methods have been proposed. Currently, the best performing methods are based on convolutional neural networks (CNNs) and require extensive datasets for training. However, at test time, they fail to impose consistency between the super-resolved image and the given low-resolution image, a property that classic reconstruction-based algorithms naturally enforce in spite of having poorer performance. Motivated by this observation, we propose a new framework that joins both approaches and produces images with superior quality than any of the prior methods. Although our framework requires additional computation, our experiments on Set5, Set14, and BSD100 show that it systematically produces images with better peak signal to noise ratio (PSNR) and structural similarity (SSIM) than the current state-of-the-art CNN architectures for SR. 

\end{abstract}

\section{Introduction}
\label{sec:intro}

Image sensing technology is fundamentally limited by the physics of the
acquisition process and often produces images with resolution below
expectation. Super-resolution (SR) techniques overcome this by taking as input
a low-resolution (LR) image and by outputting a high-resolution (HR) version of
it. This, however, requires filling in the missing entries of the image, for
which there are many possibilities. Predicting these entries thus requires
making assumptions about the images, and different techniques make different
assumptions. Landmark techniques include polynomial interpolation methods,
which assume that images are spatially smooth and well approximated by bilinear
or bicubic functions~\cite{Keys81-CubicConvolutionInterpolationForDigitalImageProcessing,Interpolation}, reconstruction methods, which assume that
images have sparse representations in certain domains that are
fixed~\cite{Fattal,ROMP,gradient} or that can be learned from training
data~\cite{Yang10-ImageSuperResolutionViaSparseRepresentation,Zeyde12-OnSingleImageScaleUpUsingSparseRepresentations}, and convolutional neural network (CNN)
architectures~\cite{SRCNN,DRCN,SRResNet}, which assume that image patches share high-level features that can be learned by a CNN. 

Indeed, the best performing SR methods are based on CNNs~\cite{SRCNN,DRCN,SRResNet}.
Although they produce outstanding results, CNNs have well-known shortcomings:
they require extensive data and computational resources for training, they have
limited theoretical guarantees and, when the training data is not
representative, they may fail to generalize. Reconstruction methods, on the
other hand, typically super-resolve images via optimization algorithms, which
confers them more interpretability and equips them with theoretical guarantees.
They use regularizers to directly encode prior knowledge about images, e.g.,
that they have relatively few edges, sparse representations in some domain, or
many recurring patterns. Although this direct encoding of prior knowledge
dispenses reconstruction-based methods from a training stage, they are
outperformed by learning-based methods.

\mypar{Our framework} 
We propose a framework that joins reconstruction and learning techniques. Our motivating observation is that the best learning techniques, namely CNN architectures, ignore important information at test time: they fail to guarantee that the given input image matches their output when downsampled. Reconstruction algorithms, on the other hand, impose this constraint naturally. Our main idea is then to integrate the HR output of a learning-based method into a reconstruction scheme, i.e., we use an image super-resolved
by a learning-based method as side information in a reconstruction-based method.
For the latter, we adopt a SR algorithm based on total variation (TV)
minimization, a prior that encodes the fact that images have a small number of
edges. This simple prior is enough to illustrate the benefits of our framework
which, nevertheless, can be adapted to more complex priors.  

Our framework introduces a
simple post-processing step that requires some additional computation. However,
as our experiments show, it systematically improves the outputs of several CNN
architectures, producing images with better peak signal to noise ratio (PSNR)
and structural similarity (SSIM)~\cite{ssim}.

\section{Related Work}
\label{Sec:RelatedWork}

Single-image SR algorithms can be split into interpolation,
reconstruction, and learning methods. 
We focus our overview on reconstruction and learning methods, as both outperform interpolation-based schemes. 

\subsection{Reconstruction-based SR}
\label{SubSec:RelW-ReconstructionBasedSR}

Reconstruction-based methods super-resolve images with an algorithm
that solves an optimization problem usually containing two terms: one
ensuring that the downsampled solution coincides with the input
LR image and another encoding prior knowledge about images. 

\mypar{Statistical models}
The work in~\cite{Fattal} models the gradient profile of an image as a Gauss-Markov random field and performs SR via maximum likelihood estimation with the constraint that downsampled solution coincides with the given LR image.
In~\cite{gradient}, images are super-resolved by ensuring that the gradient of
the HR image is close to a gradient profile obtained from the LR image using a
similar, but parametric model, while imposing again that the downsampled
solution coincides with the given LR image.

\mypar{Total variation methods}
A different line of work directly encodes prior knowledge into the optimization
problem. For instance,
\cite{Rudin92-NonlinearTotalVariationBasedNoiseRemovalAlgorithms}
introduced the concept of TV to capture the number of edges
in an image and observed that natural images have small TV. Since then, several
algorithms have been proposed to super-resolve images by minimizing TV. For example, \cite{Morse01-ImageMagnificationUsingLevelSetReconstruction} addresses the problem by discretizing a differential equation that
relates changes in the values of individual pixels to the curvature of level
sets. More recently, with the development of
nondifferentiable convex optimization methods, TV minimization has been
addressed in the discrete setting by a wide range of
algorithms~\cite{TV2D,BioucasDias07-ANewTwIST,Goldstein09-TheSplitBregmanMethodForL1RegularizedProblems,Becker11-NESTA,TVAL3}.
There are essentially two versions of discrete TV, according to the norm that
is applied to the discretized gradient at each pixel: isotropic ($\ell_2$-norm)
and anisotropic ($\ell_1$-norm); see, e.g.,
\cite{BioucasDias07-ANewTwIST,TVAL3}. Since both versions are convex, discrete
TV minimization algorithms not only have small computational complexity (as
matrix-vector operations can be performed via the FFT algorithm), but are also
guaranteed to find a global minimizer.

\mypar{Alternative regularizers}
Natural images also have simple representations in other domains, and
other regularizers have been used for SR. For
example, \cite{ROMP} uses the fact that images have sparse wavelet
representations. The
patches of natural images also tend to lie on a low-dimensional
manifold~\cite{Lee03-TheNonlinearStatisticsOfHighContrastPatchedInNaturalImages}, which has motivated nonlocal methods. These use as
regularizers (nonconvex) functions that enforce many patches of an image to have similar
features; see, e.g.,
\cite{Peyre08-NonlocalRegularizationOfInverseProblems,Protter09-GeneralizingTheNonlocalMeansToSuperResolutionReconstruction}. 
Finally, different regularizers can be combined as
in~\cite{Shi15-LRTV,Wu16-ASelfLearningImageSuperResolutionMethodViaSparseRepresentationAndNonLocalSimilarity}.

\subsection{Learning-based SR}
 Learning-based methods usually operate on patches and learn how to map
a LR to a HR patch. They consist of a training stage, in which the map is
learned, and a testing stage, in which the map is applied to the LR patches to
obtain HR patches.

\mypar{Coding and dictionary learning} 
Coding and dictionary learning algorithms super-resolve the patches of an image by linearly combining HR patches collected (or created) from other images during training. An early
method is~\cite{NeighborEmbedding}, which adapts the locally linear embedding
algorithm of~\cite{LLE} for SR. Drawing from developments in sparse-based
reconstruction, \cite{Yang10-ImageSuperResolutionViaSparseRepresentation}
proposed using training images to learn two dictionaries, one for HR and
another for LR patches, such that corresponding HR and LR patches have the same
coefficients; see~\cite{Zeyde12-OnSingleImageScaleUpUsingSparseRepresentations}
for a similar approach. While these schemes rely on other images to learn
the LR-HR map, \cite{selfsim,SelfExSR} use self-similarity to learn the map without using other images. 

\mypar{Regression methods} 
Regression methods compute the LR-HR map using a set of basis
functions whose coefficients are computed via regression. The key idea is to
perform clustering on the training images, and use algorithms like
kernel ridge regression (KRR)~\cite{Kim}, Gaussian process
regression~\cite{gaussian} and random forests~\cite{RandomForest} to perform SR.

\mypar{CNN-based} 
Currently, the best performing SR methods are based on CNNs.
Taking inspiration in the dictionary learning algorithm
in~\cite{Yang10-ImageSuperResolutionViaSparseRepresentation},
SRCNN~\cite{SRCNN} was one of the first CNNs performing image SR. It consists
of a patch extraction layer, a representation layer of the non-linear mappings,
and a final layer that outputs the reconstructed image.  As most neural
network architectures, SRCNN requires an extensive training set, and has to be
retrained for each different upscaling factor. To overcome the latter problem, DRCN~\cite{DRCN} uses a recursive convolutional
layer that is repeatedly applied to obtain SR. Building on generative
adversarial networks~\cite{GAN}, state-of-the-art SR results are
obtained in~\cite{SRResNet}. It proposes two networks: SRGAN, which produces
photo-realistic images with high perceptual quality, and SRResNet, which
achieves the best PSNR and SSIM compared to all prior SR approaches. Although CNN-based methods achieve state-of-the-art SR results, they suffer from
a major shortcoming that leaves room for improvement: in general, they fail to
impose consistency between the HR and LR image in the testing stage. Such
consistency, however, is always enforced by reconstruction-based (specifically,
optimization-based) algorithms. This observation is the main motivation behind
our framework, which we introduce next.

 \begin{figure}[t]
	\centering
\includegraphics[scale=0.2]{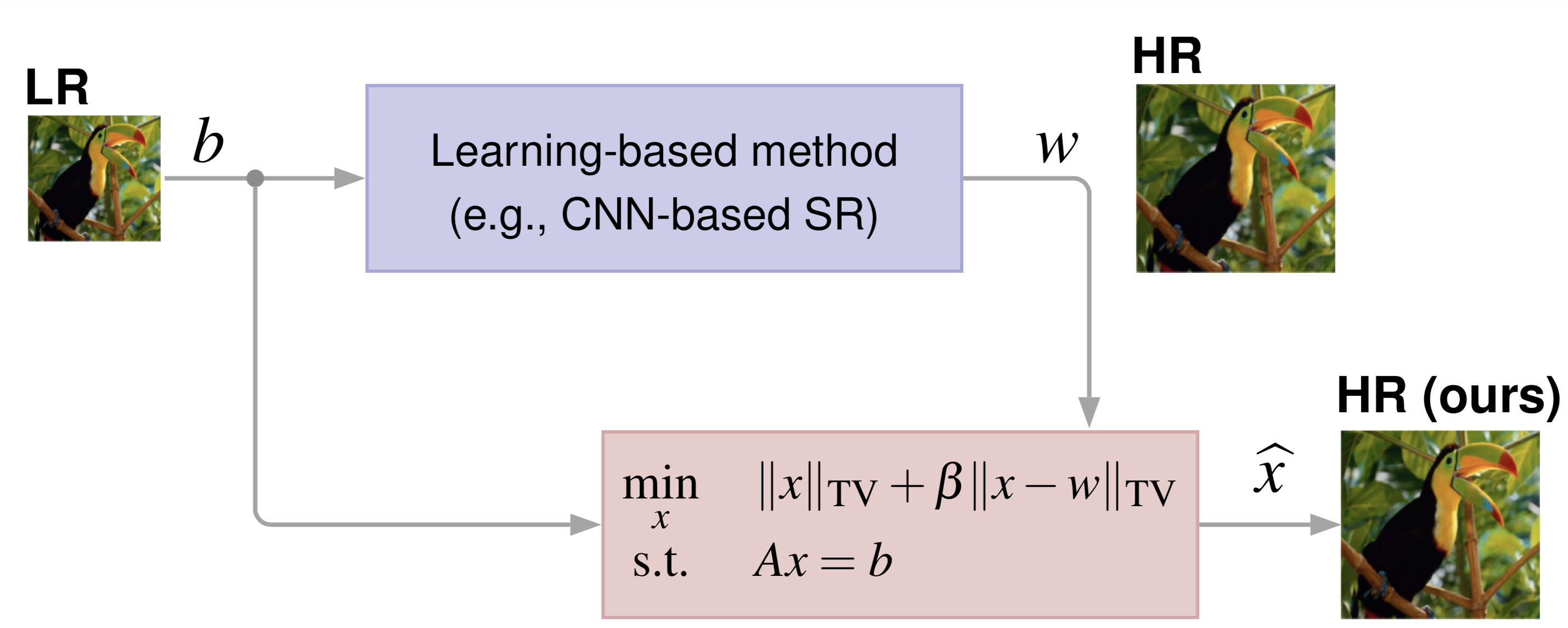}
  \caption{Our framework: a learning-based method super-resolves $b$ into $w$,
    and both images are then used to find the final estimate $\widehat{x}$ of a TV-TV minimization problem.}
  \label{Fig:BlockDiagram}
\end{figure}                        

\section{Our Framework}
\label{Sec:3}

Given a low-resolution (LR) image, our goal is to build a high-resolution (HR)
version of it. Before describing how we achieve this, we introduce our model
and assumptions.

\mypar{Model and assumptions}
We denote by $X^\star \in \mathbb{R}^{M \times N}$ the HR image, and by
$x^\star := \text{vec}(X^\star)\in \mathbb{R}^n$ its column-major
vectorization, where $n := M\cdot N$. Since these represent 2D quantities, we will always work either with grayscale images or with specific channels of
a given representation of color images. We assume that $X^\star$ has a small number of edges. Namely, we assume
$\|x^\star\|_{\text{TV}} := \sum_{i=1}^{M}\sum_{j=1}^{N} |D_{ij}^v x^\star| +
|D_{ij}^h x^\star|$ is small, where $D_{ij}^v$ (resp.\ $D_{ij}^h$) is a
row-vector that extracts the vertical (resp.\ horizontal) difference at pixel
$(i,j)$ of $X^\star$. Notice that this corresponds to the anisotropic
TV-norm~\cite{BioucasDias07-ANewTwIST,TVAL3}, which can be written more
succinctly as $\|x^\star\|_{\text{TV}} = \|Dx^\star\|_1$, where the rows of $D
\in \mathbb{R}^{2n \times n}$ contain all the $D_{ij}^v$'s and $D_{ij}^h$'s. We
assume periodic boundary conditions, so that matrix-vector products can be
efficiently computed via the FFT algorithm. 

The vectorization of the given LR
image will be denoted by $b \in \mathbb{R}^m$, where $m < n$. We assume the LR and HR vectorizations are
linearly related as $b = Ax^{\star}$, where $A\in \mathbb{R}^{m\times n}$ is a downsampling operator, i.e., $Ax^{\star}$ represents a downsampled version of $x^{\star}$, which can be obtained via filtering (e.g., bicubic or box) or even direct sampling. As bicubic filtering is a popular downsampling operator, and is the preferred method for generating training data for CNNs, we will assume that A implements such (linear) operator.

\mypar{Proposed framework}
Fig.~\ref{Fig:BlockDiagram} shows the main components of our framework. It
builds on a \textit{base method} which, because of their current
state-of-the-art performance, will be a learning-based scheme. The LR image $b$
is fed into the base method, which super-resolves it into a HR image
that we will denote by $w \in \mathbb{R}^n$. As mentioned in
Section~\ref{Sec:RelatedWork}, learning-based schemes, namely the ones based on
CNN architectures, fail to enforce consistency between the LR and HR images during testing. That is, in our notation, fail to guarantee $Ax^\star = b$.
As shown in Fig.~\ref{Fig:BlockDiagram}, we thus propose an additional
stage, which processes both the given LR image $b$ and the HR image $w$ from
the learning-based method to build our final estimate $\widehat{x}$ of $x^\star$. That stage
solves a problem that we call TV-TV minimization.

\mypar{TV-TV minimization}
Inspired by the success of sparse reconstruction schemes that integrate prior
information in the form of past, similar
examples~\cite{Chen08-PriorImageConstrainedCS,Weizman16-ReferenceBasedMRI,Mota17-CompressedSensingWithPriorInformation},
we propose to estimate $x^\star$ by solving an optimization problem whose
objective balances the TV-norm of the optimization variable $x$ and the TV-norm
of the difference between $x$ and $w$, while constraining $Ax = b$. That is, we
obtain the final estimate $\widehat{x}\,$ by solving \textit{TV-TV
minimization}:
\begin{equation}
  \label{Eqn:TVTV}
  \begin{array}[t]{ll}
    \underset{x}{\text{minimize}} & \|x\|_{\text{TV}} +  \beta \|x-w\|_{\text{TV}}\\
    \text{subject to} &  Ax = b\,,
  \end{array}
\end{equation}
where $\beta \geq 0$ is a tradeoff parameter. Since the constraints
of~\eqref{Eqn:TVTV} are linear and the TV-norm is convex, problem~\eqref{Eqn:TVTV}
is convex. To our knowledge, the first instance of~\eqref{Eqn:TVTV} appeared
in~\cite{Chen08-PriorImageConstrainedCS} in the context of dynamic CT imaging,
where $w$ was a coarse estimation of the image to be reconstructed. The work
in~\cite{Weizman16-ReferenceBasedMRI} generalized that approach, in the context
of MRI imaging, to the case where the TV-norms are weighted, and their weights
are updated as in~\cite{Candes08-EnhancingSparsityReweightedL1Minimization}.
There, $w$ represented a reference image, i.e., an image similar to the image to reconstruct.

Problem~\eqref{Eqn:TVTV} enforces consistency between the
LR image $b$ and the final estimate $\widehat{x}$ via the constraint $Ax =
b$. Among the possible solutions of $Ax=b$, it seeks the one that has a small TV-norm and, at the same time, does not deviate much
from $w$, in the sense that their discrete gradients are similar. While it is
also possible to enforce pixel-level similarity, we found that imposing
similarity of gradients produces the best results.

As mentioned in Section~\ref{SubSec:RelW-ReconstructionBasedSR}, it is possible
to use alternative regularizers that capture other properties of natural
images, or even combine several different regularizers. Indeed, the reason why
we are proposing a ``framework'' and not just a ``method'' is because
we allow both blocks in Fig.~\ref{Fig:BlockDiagram} to vary. Using TV-TV
minimization~\eqref{Eqn:TVTV}, however, enables us to draw from the insights
in~\cite{Mota17-CompressedSensingWithPriorInformation} for $\ell_1$-$\ell_1$
minimization.

\mypar{Connections with \boldmath{$\ell_1$}-\boldmath{$\ell_1$} minimization}
Since $\|x\|_{\text{TV}} = \|Dx\|_1$, by introducing an additional variable $
\mathbb{R}^{2n} \ni z = Dx$, we rewrite~\eqref{Eqn:TVTV} as 
\begin{equation}
  \label{Eqn:TVTVintermediate}
  \begin{array}[t]{ll}
    \underset{z,x}{\text{minimize}} & \|z\|_1 +  \beta \|z-\overline{w}\|_1 \\
    \text{subject to} &  Ax = b\,,\,\,\,\,\,\, Dx = z\,,
  \end{array}
\end{equation}
where we defined $\overline{w} := Dw$. We denote the concatenation of
the optimization variables as $\overline{x} := (z, x) \in
\mathbb{R}^{3n}$ and define  $\overline{A} := \begin{bmatrix}
    0_{m\times 2n} & A;& -I_{2n} & D\end{bmatrix}$, and $\overline{b} :=
  \begin{bmatrix}
    b & 0_{2n}
  \end{bmatrix}^\top$,
where $0_{a\times b}$ (resp.\ $0_{a}$) stands for the zero matrix (resp.\
vector) of dimensions $a \times b$ (resp.\ $a \times 1$), and $I_{2n}$ is the
identity matrix in $\mathbb{R}^{2n}$. This enables us to further
rewrite~\eqref{Eqn:TVTVintermediate} as
\begin{equation}
  \label{Eqn:L1L1}
  \begin{array}[t]{ll}
    \underset{\overline{x}}{\text{minimize}} & \|P_{2n} \overline{x}\|_1 
    + \beta \|P_{2n} \overline{x} - \overline{w}\|_1\\
    \text{subject to} &  \overline{A}\overline{x} = \overline{b}\,,
  \end{array}
\end{equation}
where $P_{2n} \in \mathbb{R}^{2n \times 3n}$ contains the first $2n$ rows of
the identity matrix in $\mathbb{R}^{3n}$, $I_{3n}$. That is, for a vector $v
\in \mathbb{R}^{3n}$, $P_{2n} v$ extracts the first $2n$ components of $v$.
The work in~\cite{Mota17-CompressedSensingWithPriorInformation} characterized
the performance of the problem in~\eqref{Eqn:L1L1} when $P_{2n}$ is the full
identity matrix, rather than just part of it, a problem they called
$\ell_1$-$\ell_1$ minimization. Assuming that $\overline{A}$ has Gaussian
random entries, \cite{Mota17-CompressedSensingWithPriorInformation} provided
precise reconstruction guarantees for such a problem. In particular, they
showed that $\beta = 1$ yields the best performance in theory and in practice. Although these results were proved only for the case in which $\overline{A}$ is
a Gaussian matrix, experiments
in~\cite{Mota16-AdaptiveRateReconstructionOfTimeVaryingSignals} suggest that
they also hold for other types of matrices. This led us to select $\beta = 1$ in all our experiments.

\mypar{Solving~\eqref{Eqn:TVTV}}
A disadvantage of our framework compared to learning-based methods, in
particular CNN architectures, is that solving an optimization problem
like~\eqref{Eqn:TVTV} requires some computation. Yet, because the downsampling
matrix $A$ and difference matrix $D$ are very structured, we can design
algorithms that take advantage of fast matrix-vector multiplications.
Specifically, given $u \in \mathbb{R}^n$ and $v \in \mathbb{R}^{m}$, $Au$
outputs a subvector of $u$, and $A^\top v$ outputs an $n$-dimensional vector
whose entries are the entries of $v$ (at locations specified by $A$) or zeros.
These operations require neither explicit construction of $A$, nor any
floating-point operation (just memory access). Because we assume periodic
boundary conditions, the matrix-vector products $Du$ and $D^\top v$, for $u \in
\mathbb{R}^n$ and $v \in \mathbb{R}^{2n}$, can be efficiently computed using
the FFT algorithm, which requires $O(n\log n)$ floating-point operations;
see~\cite{TVAL3}. 

To solve~\eqref{Eqn:TVTV}, we apply ADMM~\cite{Boyd11-ADMM}, guaranteeing
that its implementation uses only matrix-vector products as above. ADMM is
applied directly to problem~\eqref{Eqn:TVTVintermediate} with the
splitting $f(z,x) = \|z\|_1 + \beta\|z - w\|_1$ and $g(x,z) = \text{i}_{\{(z,
x)\,:\, Ax = b,\, Dx = z\}}(z, x)$, where $\text{i}_S(u)$ is an indicator
function, i.e., $\text{i}_S(u) = 0$ if $u \in S$, and
$\text{i}_S(u) = +\infty$ otherwise. Our focus was not on obtaining the most
efficient implementation, but we point out possible improvements in
Section~\ref{Sec:Conclusions}. 

\section{Experiments}
\label{Sec:4}

We tested our framework on the standard test sets Set5~\cite{Set5}, Set14~\cite{Zeyde12-OnSingleImageScaleUpUsingSparseRepresentations}, and BSD100 \cite{BSD100}, each containing respectively 5, 14, and 100 images. For RGB images, we only super-resolved the luminance channel of the YCbCr color space, as in~\cite{SelfExSR,SRCNN, DRCN,SRResNet}. All our experiments were run on the same machine with a 12 core 2.10 GHz Intel(R) Xeon(R) Silver 4110 CPU and two Nvidia GeForce RTX GPUs. The code used for all experiments is available online\footnote{\url{https://github.com/marijavella/sr-tvtvsolver}}.

 We used four state-of-the art learning schemes: KRR by Kim~\cite{Kim}, self-similarity~\cite{SelfExSR}, pre-trained CNNs~\cite{SRCNN, DRCN} and basic TV minimization, i.e., \eqref{Eqn:TVTV} with $\beta=0$, using TVAL3~\cite{TVAL3}. Of these methods, only~\cite{TVAL3} operates on full images, just like our TV-TV minimization algorithm; all the remaining methods  operate only on patches. 
 The output images of SRCNN~\cite{SRCNN}, Kim~\cite{Kim} and SelfExSR~\cite{SelfExSR} were obtained from the repository\footnote{\url{https://github.com/jbhuang0604/SelfExSR}}. 

 As our framework is tied to a given base method, whenever the latter changes, so does our method. For assessing performance, we used PSNR and SSIM~\cite{ssim} computed on the luminance channel. 
 
\begin{figure}[t]
  \centering 
%
%

\includegraphics[width= 12.5cm, height=3.3cm]{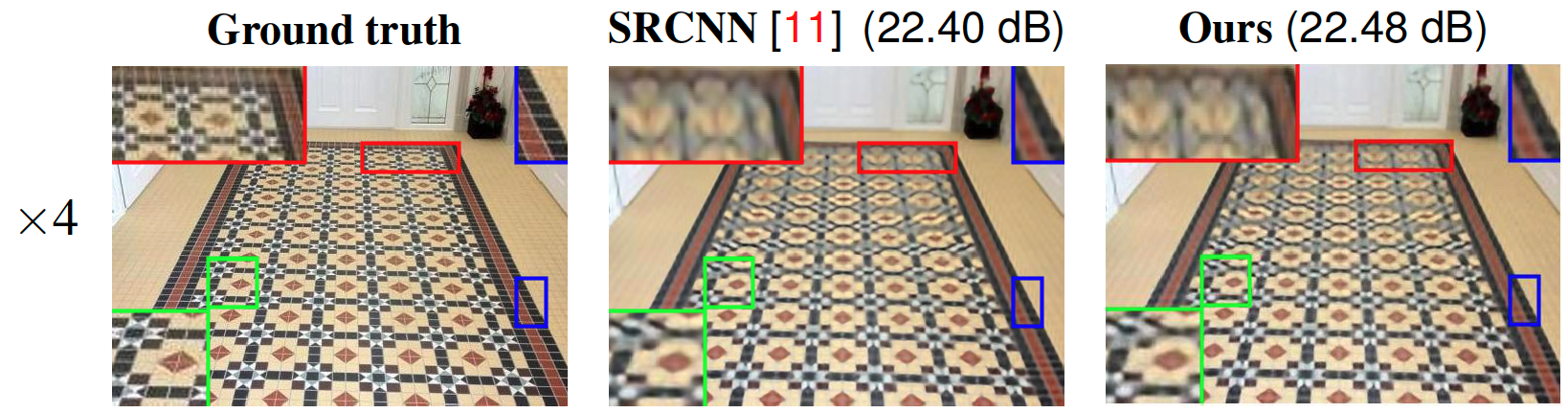}

  \caption{
    Performance of our method, using SRCNN~\cite{SRCNN} as
    the base method, for an image not included in any of the training or
    test sets. The ground truth was downsampled by $4$ and
    then super-resolved with SRCNN and our method. PSRN values are shown
    in parenthesis.
  }
  \vspace{-0.5cm}
  \label{Fig:tiles}
\end{figure}

\subsection{Illustrating consistency between LR and HR images}
\label{SubSec:ResultsConsistency}

To demonstrate how imposing consistency between the LR and HR images can
improve SR performance, we conducted an experiment using an image that is not
in any of the above datasets. The image, of dimensions 472 $\times$ 352, is shown in the left of Fig.~\ref{Fig:tiles} and was selected because of its structured patterns. We downsampled it via bicubic filtering by a factor of $4$, and then reconstructed it with the CNN architecture
SRCNN~\cite{SRCNN}, and with our method using SRCNN as the base algorithm. Since SRCNN fails to enforce consistency
between the LR and HR images during testing, its output (Fig.~\ref{Fig:tiles}, center) blurs
the tiles and removes some details. Our method, which uses the
output of SRCNN (Fig.~\ref{Fig:tiles}, center) and the $4\times$ downscaled
version of the ground truth (Fig.~\ref{Fig:tiles}), seems to reduce the blur artifacts, as manifested by an improvement in the PSNR. Such an improvement, however,
comes with a computational cost: to super-resolve that image required SRCNN 2 seconds, and solving~\eqref{Eqn:TVTV} required 151 seconds. 

\subsection{Experiments on Set4, Set5, and BSD100}

Tab.~\ref{Tab:SystematicExp} shows the results of our experiments for the testing datasets Set5, Set14, and BSD100. It displays the PSNR and SSIM values for $2\times$ and $4\times$ upscaling
tasks, and these values represent averages over all the images in the corresponding dataset. 
Recall that higher PSNR and SSIM values reflect better performance.
The table has 5 parts, one for each variation of the base method. For
example, the first 3 rows refer to the results when the method in
Kim~\cite{Kim} is the base algorithm; in the next 3 rows, the base method is
SRCNN~\cite{SRCNN}, and so on. The values for TVAL3~\cite{TVAL3} are repeated
in all parts of the table, since its input is the same image in all the cases. 

It can be observed that TVAL3, an algorithm for solving~\eqref{Eqn:TVTV} with
$\beta = 0$, yields the worst performance. All the learning-based
methods (4th column) systematically give results better than TVAL3. Our method
(last column), in turn, uses the outputs of those algorithms
and, by enforcing consistency between the LR and HR images, systematically
improves their outputs: for example, the range of the gains
in PSNR values was $0.006$-$0.177$ dB.

\begin{figure}[H]
\hspace{-0.9cm}
\includegraphics[scale=1]{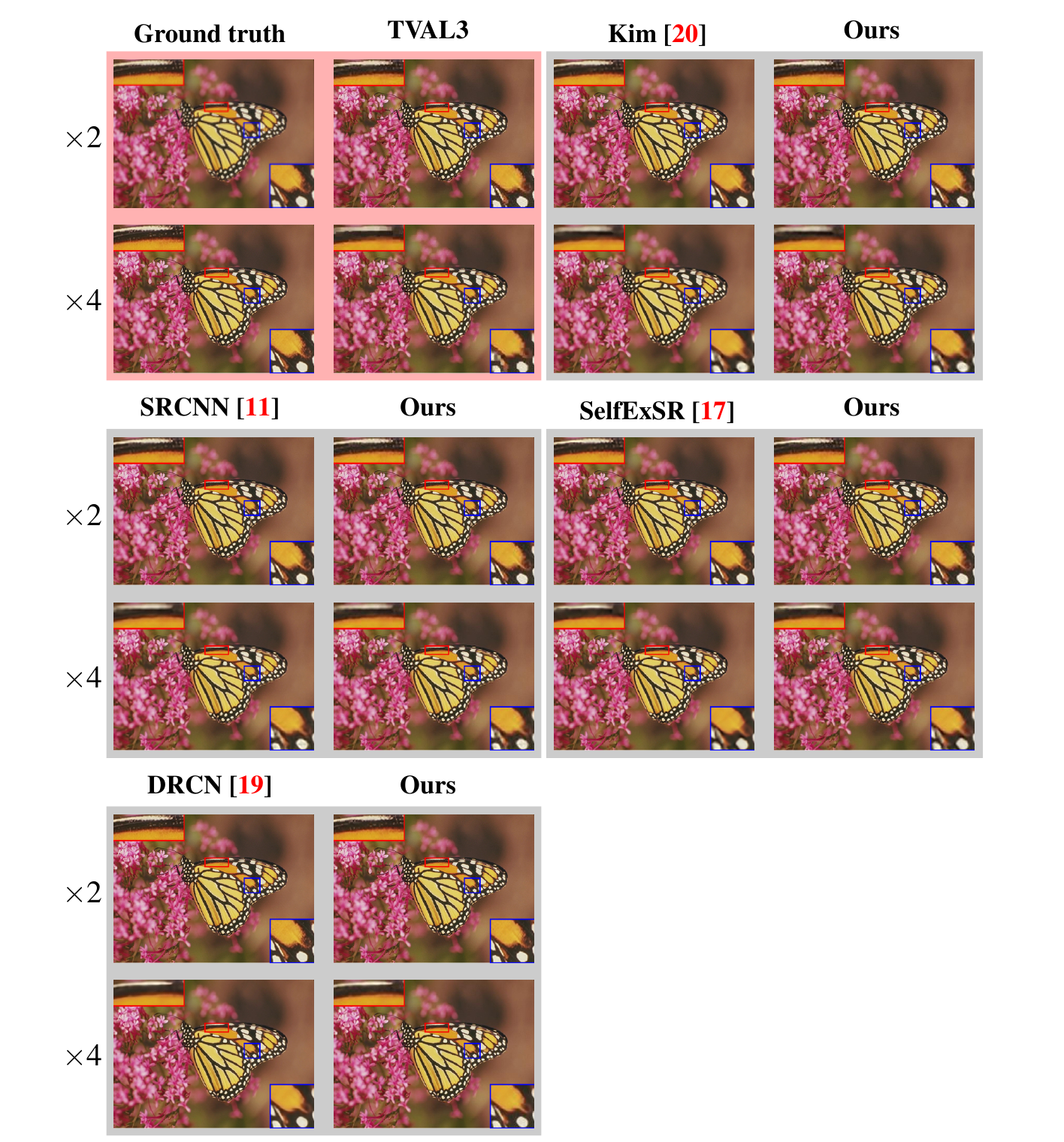}

  \caption{Reconstructions of the Monarch image (Set14) for all the methods in
  Tab.~\ref{Tab:SystematicExp}. Some details are highlighted (best viewed by
  zooming in on a computer screen).}
  \label{Fig:Set14ex}
\end{figure}

Fig.~\ref{Fig:Set14ex} shows the reconstructions of a particular image in
Set14 for all the methods. The top-left box corresponds to the ground truth and
TVAL3, which were the same in all the experiments. The remaining boxes
show the results of a given learning-based method and the corresponding output using our framework. As in Section~\ref{SubSec:ResultsConsistency}, it
can be seen that the learning-based methods tend to produce images that lack
details and they fail to impose consistency with the given LR image. Our
method, in contrast, constraints the LR image to be consistent with the HR image which improves the image quality and slightly reduces the blurring effect. The execution times of SRCNN~\cite{SRCNN} and DRCN~\cite{DRCN} to super-resolve the sample image in Fig.~\ref{Fig:Set14ex} are shown in Tab.~\ref{table2}. Although our method yields better PSNR/SSIM values, it requires significantly more computational time. 

\section{Conclusions}
\label{Sec:Conclusions}

We proposed a framework for super-resolving images that leverages the good
performance of learning-based methods, including CNN architectures, and the
consistency enforced by reconstruction-based algorithms. The framework takes a
base learning-method, and improves its output by solving a TV-TV minimization
problem. Although the image model we used is quite simple, it suffices to
illustrate the reconstruction gains in PSNR and SSIM that can be obtained. Our
experimental results involving several state-of-the-art SR algorithms show that
the proposed framework achieves some performance gains over those
algorithms, however, at an additional computation cost. 

Possible research directions include using regularizers that capture
more realistic image assumptions and accelerating the algorithm for solving
the TV-TV minimization algorithm e.g., using a C implementation, or a
trained recurrent neural network as a solver~\cite{Gregor10-LearningFastApproximationsOfSparseCoding}.

\begin{table}[H]
	\centering
	\renewcommand{\arraystretch}{1.1}
	\caption{Average PSNR (SSIM) results using the reference methods.}
	\label{Tab:SystematicExp}
	\def\tbsp{0.15cm}
	\scalebox{0.97}{
		\begin{tabular}{@{}llllllll@{}}
			\toprule[1.1pt]
			\textcolor{thisblue}{\bf\normalsize Dataset}                     && 
			\textcolor{thisblue}{\bf\normalsize Scale}                       && 
			\textcolor{thisblue}{\bf\normalsize TVAL3} \cite{TVAL3}          &&
			\textcolor{darkredc}{\bf\normalsize\textit{Kim}}~\cite{Kim}      &
			\textcolor{thisblue}{\bf\normalsize Ours}
			\\
			\midrule
			\multirow{2}{*} {Set5} && $\times2$ && 34.0315 (0.9354)&&36.2465 (0.9516)  & \textbf{36.4499} (\textbf{0.9537})  \\
			&& $\times4$ && 29.1708 (0.8349) && 30.0730 (0.8553)& \textbf{30.2289} (\textbf{0.8593})  \\
			\midrule
			\multirow{2}{*}{Set14} && $\times2$ && 31.0033 (0.8871) && 32.1359 (0.9031) & \textbf{32.3044} (\textbf{0.9055})\\
			&& $\times4$ && 26.6742 (0.7278)  &&  27.1836 (0.7434)  & \textbf{27.2993} (\textbf{0.7488}) \\
			\midrule
			\multirow{2}{*}{BSD100} && $\times2$ && 30.1373 (0.8671) && 31.1124 (0.8840)  & \textbf{31.2097} (\textbf{0.8864}) \\
			&& $\times4$ && 26.3402 (0.6900) &&26.7099  (0.7027)  & \textbf{26.7891} (\textbf{0.7086}) \\[\tbsp]
			\midrule[1pt]
			\textcolor{thisblue}{\bf\normalsize Dataset}                     && 
			\textcolor{thisblue}{\bf\normalsize Scale}                       && 
			\textcolor{thisblue}{\bf\normalsize TVAL3} \cite{TVAL3}          &&  
			\textcolor{darkredc}{\bf\normalsize \textit{SRCNN}}~\cite{SRCNN} &
			\textcolor{thisblue}{\bf\normalsize Ours}
			\\
			\midrule
			\multirow{2}{*}{Set5} && $\times2$ && 34.0315 (0.9354) &&  36.2772 (0.9509) & \textbf{36.5288} (\textbf{0.9536})   \\
			&& $\times4$ &&  29.1708 (0.8349) && 30.0765 (0.8525)  & \textbf{30.2669} (\textbf{0.8590}) \\
			\midrule
			\multirow{2}{*}{Set14} && $\times2$ && 31.0033 (0.8871)&& 31.9954 (0.9012)   & \textbf{32.2949} (\textbf{0.9057}) \\
			&& $\times4$ && 26.6742 (0.7278)   && 27.1254 (0.7395) &\textbf{27.3040} (\textbf{0.7480}) \\
			\midrule
			\multirow{2}{*}{BSD100} && $\times2$ && 30.1373 (0.8671) &&  31.1087 (0.8835) & \textbf{31.2241} (\textbf{0.8866}) \\
			&& $\times4$ && 26.3402 (0.6900) && 26.7027 (0.7018)&\textbf{26.7838} (\textbf{0.7085}) \\[\tbsp]
			
			\midrule[1.1pt]
			\textcolor{thisblue}{\bf\normalsize Dataset} && 
			\textcolor{thisblue}{\bf\normalsize Scale} && 
			\textcolor{thisblue}{\bf\normalsize TVAL3} \cite{TVAL3}  &&  
			\textcolor{darkredc}{\bf\normalsize \textit{SelfExSR}}~\cite{SelfExSR} &
			\textcolor{thisblue}{\bf\normalsize Ours}
			\\
			\midrule
			\multirow{2}{*}{Set5} && $\times2$ && 34.0315 (0.9354) && 36.5001   (0.9537)  & \textbf{36.5321} (\textbf{0.9542})    \\
			&& $\times4$ &&  29.1708 (0.8349) && 30.3317 (0.8623)  & \textbf{30.3370} (\textbf{0.8625})\\
			\midrule
			\multirow{2}{*}{Set14} && $\times2$ && 31.0033 (0.8871)   &&   32.2272 (0.9036)   & \textbf{32.3951} (\textbf{0.9059}) \\
			&& $\times4$ && 26.6742 (0.7278)   && 27.4014  (0.7518)   &\textbf{27.4730} (\textbf{ 0.7536}) \\
			\midrule
			\multirow{2}{*}{BSD100} && $\times2$ && 30.1373 (0.8671) &&  31.1833 (0.8855)   & \textbf{31.2056} (\textbf{0.8862}) \\
			&& $\times4$ && 26.3402 (0.6900) && 26.8459 (\textbf{0.7108})&\textbf{26.8482} (\textbf{0.7108}) \\[\tbsp]
			
			\midrule[1.1pt]
			\textcolor{thisblue}{\bf\normalsize Dataset} && 
			\textcolor{thisblue}{\bf\normalsize Scale} && 
			\textcolor{thisblue}{\bf\normalsize TVAL3} \cite{TVAL3} &&  
			\textcolor{darkredc}{\bf\normalsize \textit{DRCN}}~\cite{DRCN} &
			\textcolor{thisblue}{\bf\normalsize Ours}
			\\
			\midrule
			\multirow{2}{*}{Set5}&& $\times2$ && 34.0315 (0.9354) && 37.6279 (0.9588) & \textbf{37.6712} (\textbf{0.9591})\\
			&& $\times4$ && 29.1708 (0.8349) && 31.5344 (0.8854) &   \textbf{31.5701} (\textbf{0.8858})\\
			\midrule
			\multirow{2}{*}{Set14} && $\times2$ && 31.0033 (0.8871)    &&  33.0585 (0.9121) &\textbf{33.1038} (\textbf{0.9129)}\\
			&& $\times4$ &&  26.6742 (0.7278)  &&  28.0269 (0.7673)  &\textbf{28.0588} (\textbf{0.7680})\\
			\midrule
			\multirow{2}{*}{BSD100} && $\times2$ && 30.1373 (0.8671)&& 31.8536 (0.8942) & \textbf{31.8737} (\textbf{0.8953}) \\
			&& $\times4$ && 26.3402 (0.6900) &&  27.2364 (0.7233) & \textbf{27.2524} (\textbf{0.7240}) \\[\tbsp]
			
			\bottomrule[1.1pt]
			
		\end{tabular}
	}
\end{table}

\vspace{-0.8cm}
\begin{table}[H]
	\centering
	\renewcommand{\arraystretch}{1.1}
	\caption{Execution times (in secs) for reconstructing the image in Fig. \ref{Fig:Set14ex}.}
	\label{table2}
	\def\tbsp{0.15cm}
	\scalebox{0.90}{
		\begin{tabular}{@{}lllllllllll@{}}
			\toprule[1.1pt]
			&& && 
			\multicolumn{3}{c}{
				\textcolor{thisblue}{\bf\normalsize Base:}
				\textcolor{darkredc}{\bf\normalsize\textit{SRCNN}}~\cite{SRCNN}
			}
			&&
			\multicolumn{3}{c}{
				\textcolor{thisblue}{\bf\normalsize Base:}
				\textcolor{darkredc}{\bf\normalsize\textit{DRCN}}~\cite{DRCN}
			}
			\\ 
			\cmidrule(lr){5-7}
			\cmidrule(lr){9-11}
			\textcolor{thisblue}{\bf\normalsize Scale}                       && 
			\textcolor{thisblue}{\bf\normalsize TVAL3} \cite{TVAL3}          &&
			\textcolor{thisblue}{\bf\normalsize SRCNN}  &&
			\textcolor{thisblue}{\bf\normalsize Ours}
			&&
			\textcolor{thisblue}{\bf\normalsize DRCN}
			&&
			\textcolor{thisblue}{\bf\normalsize Ours}
			\\
			\midrule
			$\times2$ && \textbf{3.34} && 9.31 && 307.84 && 156.20&&  306.61\\
			$\times4$ && \textbf{6.08} &&  9.43 && 279.89 && 143.20 && 299.39  \\
			\bottomrule[1.1pt]
		\end{tabular}
	}
\end{table}

\pagebreak
\section{Acknowledgments}
Work supported by EPSRC EP/S000631/1 via UDRC/MoD.
\bibliography{References}
\end{document}